%% file: main.tex
\documentclass[lettersize,journal]{IEEEtran}
\usepackage{amsmath,amsfonts}
\usepackage{algorithmic}
\usepackage{algorithm}
\usepackage{array}
\usepackage[caption=false,font=normalsize,labelfont=sf,textfont=sf]{subfig}
\usepackage{textcomp}
\usepackage{stfloats}
\usepackage{url}
\usepackage{verbatim}
\usepackage{graphicx}
\usepackage{amssymb}      % 提供高级数学符号（如 \mathbb, \mathcal, \nrightarrow）
\usepackage{amsfonts}    % 扩展数学字体（如黑体 \mathbb{R}）
\usepackage{amsmath}      % 基础数学环境（如 equation*, align, cases, matrix）
\usepackage{tabularray}          % 现代化表格工具（支持自动换行、单元格样式等）
\UseTblrLibrary{booktabs}        % 为 tabularray 添加 booktabs 的优雅表格线（\toprule等）
\usepackage{array, booktabs}      % 基础表格控制 + 专业表格线（与 tabularray 互补）
\usepackage{makecell}             % 单元格内换行（\makecell{第一行\\第二行}）
\usepackage[table]{xcolor}        % 表格着色（\rowcolor{gray!20}）
\usepackage{hhline}              % 绘制复杂的表格横线（解决颜色与线框冲突）
\usepackage[square,numbers,sort&compress]{natbib}

\usepackage{hyperref}
\hypersetup{
	colorlinks=true,
	linkcolor=blue,
	citecolor=blue,
	urlcolor=blue,
	breaklinks=true
}

\begin{document}

\title{RAFT-MSF++: Temporal Geometry-Motion Feature Fusion for Self-Supervised Monocular Scene Flow}

%\author{IEEE Publication Technology,~\IEEEmembership{Staff,~IEEE,}
%        % <-this % stops a space
%\thanks{This paper was produced by the IEEE Publication Technology Group. They are in Piscataway, NJ.}% <-this % stops a space
%\thanks{Manuscript received April 19, 2021; revised August 16, 2021.}
%}

\author{%
	Xunpei~Sun$^{1}$, 
	Zuoxun~Hou$^{2}$, 
	Yi~Chang$^{3}$, 
	Gang~Chen$^{1}$, 
	and~Wei-Shi~Zheng$^{1}$%
	\thanks{$^{1}$School of Computer Science and Engineering, Sun Yat-sen University, Guangzhou, China 510006.
		E-mail: sunxp7@mail2.sysu.edu.cn; cheng83@mail.sysu.edu.cn; wszheng@ieee.org.}%
	\thanks{$^{2}$Beijing Institute of Space Mechanics and Electricity, Beijing, China 100094.
		E-mail: houzx\_bisme@spacechina.com.}%
	\thanks{$^{3}$School of Artificial Intelligence and Automation, Huazhong University of Science and Technology, Wuhan, China 430074.
		E-mail: yichang@hust.edu.cn.}%
%	\thanks{*Corresponding author: Gang Chen.}%
}

% arXiv version: remove IEEE journal header and copyright notice.

\maketitle

\begin{abstract}
Monocular scene flow estimation aims to recover dense 3D motion from image sequences, yet most existing methods are limited to two-frame inputs, restricting temporal modeling and robustness to occlusions.  
We propose \textbf{RAFT-MSF++}, a self-supervised multi-frame framework that recurrently fuses temporal features to jointly estimate depth and scene flow. Central to our approach is the Geometry-Motion Feature (GMF), which compactly encodes coupled motion and geometry cues and is iteratively updated for effective temporal reasoning. 
To ensure the robustness of this temporal fusion against occlusions, we incorporate relative positional attention to inject spatial priors and an occlusion regularization module to propagate reliable motion from visible regions. These components enable the GMF to effectively propagate information even in ambiguous areas. 
Extensive experiments show that RAFT-MSF++ achieves 24.14\% SF-all on the KITTI Scene Flow benchmark, with a 30.99\% improvement over the baseline and better robustness in occluded regions. 
The code is available at \url{https://github.com/sunzunyi/RAFT-MSF-PlusPlus}.
\end{abstract}


\input{sec/1_intro}

\input{sec/2_related}

\input{sec/3_approach}
\input{sec/4_experiments}

\section{Conclusion}\label{sec13}
We present RAFT-MSF++, an unsupervised multi-frame monocular scene flow framework that recurrently fuses temporal features to jointly estimate depth and scene flow. Central to our approach is the Geometry-Motion Feature (GMF), which compactly encodes coupled geometry and motion cues for effective multi-frame fusion. 
To improve the robustness of GMF-based fusion under occlusion, we further incorporate relative positional attention and occlusion regularization, enabling better utilization of spatial priors and temporal context. 
Extensive experiments on KITTI demonstrate that RAFT-MSF++ achieves competitive performance, significantly outperforming existing multi-frame baselines and establishing strong results in occluded regions.

Overall, our findings highlight the importance of joint geometry–motion modeling and recurrent temporal fusion for robust monocular scene flow estimation.


%\newpage

%\begin{thebibliography}{1}
\bibliographystyle{IEEEtran}
\footnotesize
\bibliography{sn-bibliography}% common bib file

\newpage

\input{appendix}

\vfill

\end{document}

%% file: sec/1_intro.tex
\section{Introduction}
\IEEEPARstart{S}{cene} flow estimation aims to recover dense 3D motion fields of dynamic scenes, providing essential cues for autonomous driving and robotics~\cite{vedula2005three, liu2024difflow3d}. 
While supervised methods have progressed across modalities including stereo~\cite{Ma_2019_CVPR}, RGB-D~\cite{teed2021raft}, and LiDAR~\cite{liu2019flownet3d}, they rely on expensive sensors and labor-intensive 3D ground truth, which limits their scalability. 
Consequently, self-supervised monocular scene flow has emerged as a promising approach due to its low cost and ease of data acquisition~\cite{hur2020self}.

\begin{figure}[t]
	\centering
	\includegraphics{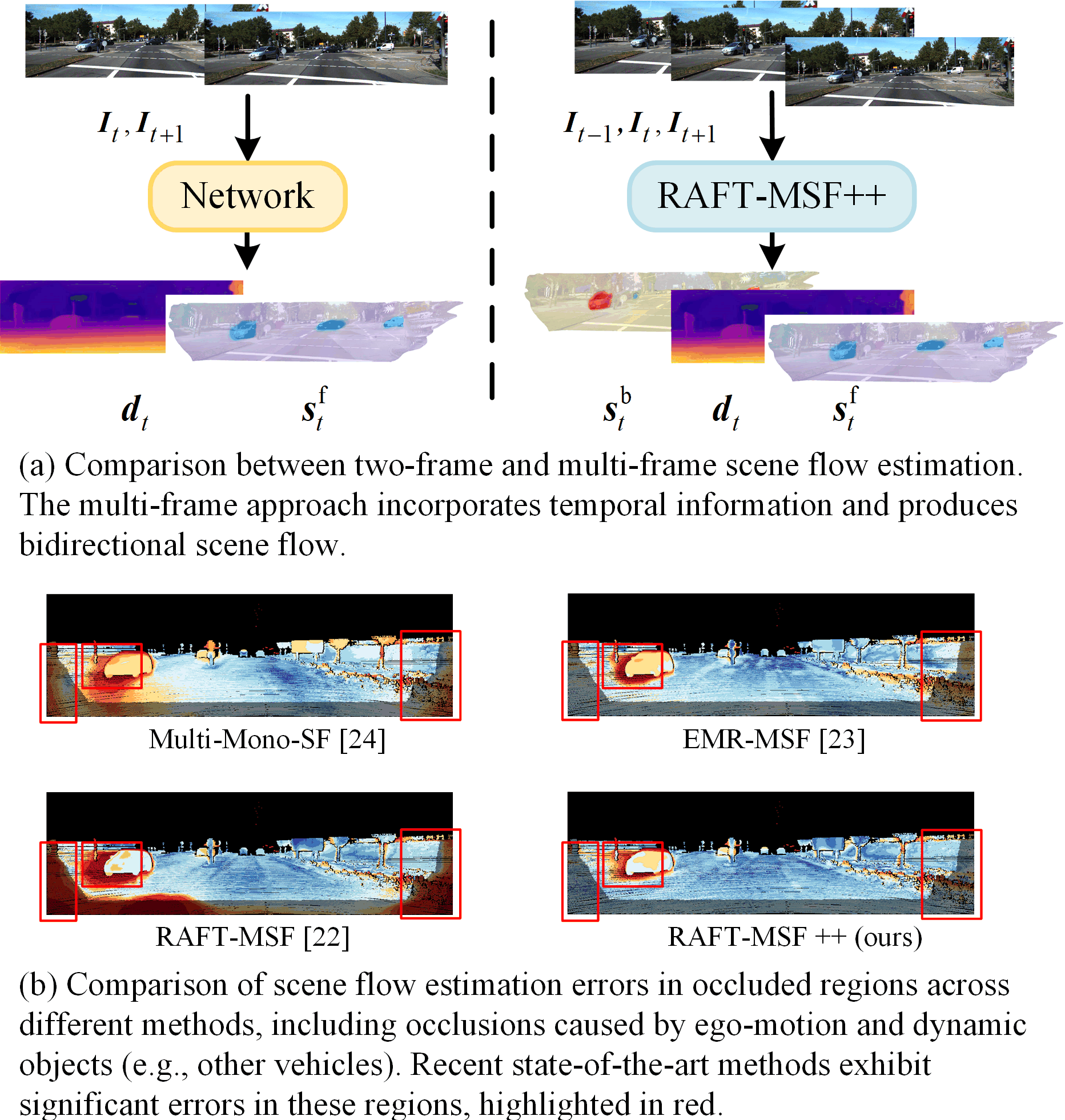}  %  [width=0.9\columnwidth]   [width=0.8\textwidth]
	\caption{
		Key challenges in monocular scene flow estimation: temporal fusion and occlusion robustness.
	}
	\label{fig:key_challenges}
\end{figure}

\begin{figure*}[h]
	\centering
	\includegraphics{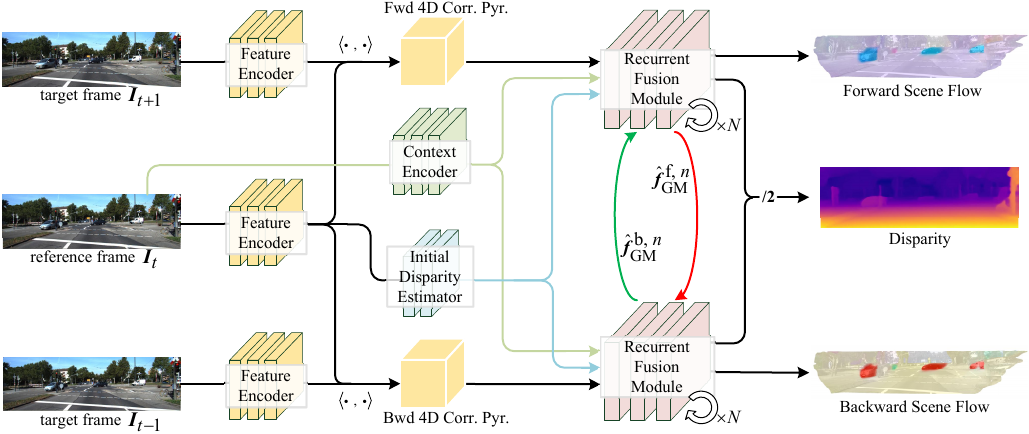}
	\caption{Proposed multi-frame architecture. The forward and backward scene flow branches run in parallel with shared weights. Disparity is derived from their average output. Instead of conventional bidirectional correlation volumes, our method uses a recurrent fusion module for temporal feature interaction.}

	\label{fig:network}
\end{figure*}

Recent efforts in end-to-end monocular scene flow have explored two-frame architectures~\cite{hur2020self, bayramli2023raft, jiang2023emr}, achieving notable accuracy and efficiency. However, two key challenges remain:

First, most state-of-the-art methods rely on two-frame inputs, overlooking temporal coherence across longer sequences, as illustrated in Fig.~\ref{fig:key_challenges}(a). While some works explore multi-frame inputs, their performance is often limited by complex bidirectional cost volumes~\cite{hur2021self} or heavy inter-frame fusion modules~\cite{mehl2023m}. These approaches lack efficient, compact mechanisms for inter-frame feature modeling, failing to fully exploit temporal information.

Second, occlusion handling remains critical in self-supervised settings. Since occluded regions violate photometric consistency, they are typically excluded from loss computation~\cite{hur2020self}, leading to inaccurate motion estimates. As shown in Fig.~\ref{fig:key_challenges}(b), effectively addressing occlusions is essential for robust scene flow estimation in dynamic environments.

To address these challenges, we propose RAFT-MSF++, a self-supervised multi-frame framework. Built upon RAFT-MSF~\cite{bayramli2023raft}, it introduces a recurrent mechanism to jointly estimate depth and scene flow by fusing inter-frame features. Central to our method is the Geometry-Motion Feature (GMF), a unified representation encoding temporal depth and motion cues. By recurrently updating and aggregating GMF across frames, the model progressively integrates temporal information, improving both expressiveness and fusion efficiency.

To ensure robustness against occlusions, we introduce two structural components. First, a relative positional attention mechanism encodes the spatial layout via learnable position embeddings, guiding GMF propagation from reliable regions to ambiguous ones through structural context. Second, an occlusion regularization module exploits motion cues from visible neighboring pixels to guide estimation in occluded regions, further improving the learning effectiveness of the GMF framework.

Extensive experiments show that RAFT-MSF++ consistently outperforms the baseline~\cite{bayramli2023raft}. On the KITTI 2015 Testing set, it reduces SF-all to 24.14\% (a 30.99\% relative improvement). 
In occluded regions (SF-occ) of the Training set, it surpasses the state-of-the-art EMR-MSF~\cite{jiang2023emr}, demonstrating superior robustness. Our contributions are:
\begin{itemize}
	
	\item We propose RAFT-MSF++, a recurrent optimizer-based framework that establishes GMF fusion as a core paradigm, effectively extending the RAFT architecture to multi-frame self-supervised monocular scene flow.
	
	\item We incorporate relative positional attention and an occlusion regularization module. These components leverage spatial priors and geometric consistency to ensure reliable GMF propagation in occluded or ambiguous regions.
	
	\item We demonstrate the effectiveness of our framework through extensive experiments, showing consistent advantages over existing methods in temporal fusion and occlusion handling.
\end{itemize}

%% file: sec/2_related.tex
\section{Related Work}
Early scene flow methods relied on energy minimization with stereo pairs~\cite{vedula2005three, vogel20153d}, while recent supervised approaches leverage explicit 3D structures from stereo, RGB-D, or LiDAR data~\cite{Ma_2019_CVPR, teed2021raft, battrawyRMSFlowNetEfficientRobust2024}. In contrast, we focus on self-supervised monocular scene flow, learning 3D motion from image sequences.

\subsection{Monocular Scene Flow}
Self-supervised monocular frameworks initially jointly predicted depth, optical flow, and camera motion~\cite{zou2018df, yin2018geonet}, but often suffered from limited accuracy. Hur and Roth~\cite{hur2020self} are the first to propose a method that directly estimates depth and 3D motion from monocular sequences. Their PWC-Net-based~\cite{sunPWCNetCNNsOptical2018} approach introduces proxy losses and joint decoders, significantly outperforming prior multi-task frameworks. Following this, RAFT-MSF~\cite{bayramli2023raft} introduces recurrent refinement, and EMR-MSF~\cite{jiang2023emr} utilizes rigid motion embeddings and multi-stage training. 
Recent extensions have focused on global feature perception~\cite{xiang2024glofp}, semantic knowledge distillation~\cite{bayramli2024integrating}, and long-range dependency modeling~\cite{chen2025mamba}. Unlike these two-frame methods, we incorporate multi-frame information to capture motion continuity and handle occlusions.

\begin{figure*}[h]
	\centering
	\includegraphics[width=1.0\linewidth]{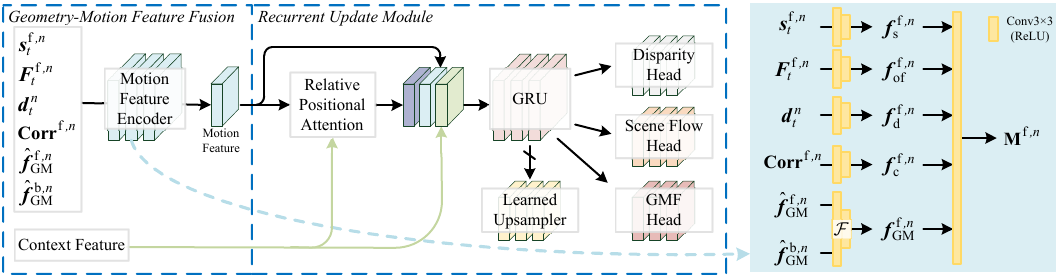}
	\caption{
		Recurrent fusion module. This figure illustrates the iterative refinement process of forward estimation. Forward and backward Geometry-Motion Features (GMFs) are integrated to enable temporal fusion. The GMFs are combined within the motion feature encoder to enhance temporal coherence. A GRU then iteratively updates the GMF, progressively refining it by incorporating contextual and motion cues. The symbol $\nrightarrow$ indicates gradient detachment.
	}
	\label{fig:update}
\end{figure*}

\subsection{Multi-Frame Estimation}
Temporal cues are vital for both optical flow and scene flow. 
In scene flow, Schuster et al.~\cite{schuster2021deep} introduce a compact motion inverter to predict forward flow from backward flow and fuse them via a convex combination. 
Multi-Mono-SF~\cite{hur2021self} uses bidirectional cost volumes and LSTMs for consistency, while M-FUSE~\cite{mehl2023m} employs a U-Net-based post-fusion module.

Research on multi-frame optical flow (the 2D projection of scene flow) is more extensive than that on scene flow. Early works explore three-frame fusion via explicit temporal modeling or feature propagation. PWC-Fusion~\cite{ren2019fusion} introduces a post-fusion module, while Back2Future~\cite{janaiUnsupervisedLearningMultiFrame2018} assumes constant velocity and enforces temporal smoothness. 
Methods such as SelFlow~\cite{liu2019selflow}, ProFlow~\cite{maurer2018proflow}, and SMURF~\cite{stone2021smurf} further enhance fusion by leveraging backward flows, backward cost volumes, or auxiliary networks. 
More recent approaches focus on recurrent or long-range modeling. RAFT~\cite{teedRAFTRecurrentAllPairs2021} employs a warm-start strategy, while recent works such as TransFlow~\cite{lu2023transflow} and VideoFlow~\cite{shiVideoFlowExploitingTemporal2023} model long-range dependencies using five-frame inputs, yet both require future frames during inference. 
VideoFlow constructs bidirectional correlation volumes in a three-frame setting and jointly predicts forward and backward flows, further extending to five frames by propagating fused features across intermediate frames. 
%Specifically, VideoFlow builds bidirectional correlation volumes in a minimal three-frame setting and jointly predicts forward and backward flows along the channel dimension. It further extends to five frames by propagating and aggregating fused bidirectional features across multiple intermediate frames. 
MemFlow~\cite{dongMemFlowOpticalFlow2024} aggregates motion features from past frames via attention mechanisms, 
while M2Flow~\cite{sun2025m2flow} models motion acceleration using four-frame inputs to learn a physically plausible motion prior.

In contrast, we propose a Geometry-Motion Feature (GMF) that directly integrates multi-frame cues into the recurrent update, avoiding explicit cost-volume fusion or post-fusion modules. 
By integrating relative positional attention and occlusion regularization to enhance this fusion, our framework achieves unified and robust multi-frame scene flow estimation.

\subsection{Occlusion Handling}
Occlusion remains a major challenge in dynamic scenes due to the lack of valid correspondences~\cite{zhang2021self, feng2024apcaflow}. Early methods handle this via masking strategies like forward-backward consistency~\cite{meister2018unflow} or distance-based checking~\cite{wang2018occlusion}, while recent optical flow works adopt distillation-based pseudo-labels~\cite{liu2019selflow, sun2025m2flow}. However, these are not directly applicable to monocular scene flow. 
In monocular scene flow, depth and 3D motion are jointly inferred from image sequences. Due to the lack of explicit depth cues, existing methods rely heavily on camera motion and scene geometry~\cite{liang2025zeroSF}.
Although artificially introducing occlusions during training (e.g., masking visible regions) can facilitate optical flow matching, such interventions tend to disrupt the spatial consistency of the image and increase uncertainty in depth prediction~\cite{hur2020self}. Consequently, most existing monocular scene flow methods~\cite{hur2020self, bayramli2023raft, jiang2023emr} exclude occluded regions from the training loss.

Unlike previous approaches, we propose an active occlusion handling strategy to improve GMF robustness, consisting of occlusion regularization and relative positional attention.

%% file: sec/3_approach.tex
\section{Method}\label{Method}
%Given three consecutive RGB frames $\{I_{t-1}, I_t, I_{t+1}\}$, we aim to estimate the 3D location $P_t$ in the reference frame $I_t$, alongside forward $s_t^f$ and backward $s_t^b$ scene flows.

Given three consecutive RGB frames $\{I_{t-1}, I_t, I_{t+1}\}$, we aim to estimate the 3D location $P_t$\footnote{Throughout this paper, $d_t^n$ and $\hat{d}_t^n$ denote disparity and depth at iteration $n$, respectively. These are related to the 3D point $P_t$ at pixel $p$ via the camera intrinsic matrix $K$ and baseline $b$ such that $\hat{d}_t^n = fb/d_t^n$ and $P_t = \hat{d}_t^n K^{-1} p$.} 
in the reference frame $I_t$, alongside forward $s_t^f$ and backward $s_t^b$ scene flows.

\subsection{Multi-frame Network Architecture}\label{Method-1}
RAFT-MSF++ extends RAFT-MSF~\cite{bayramli2023raft} by explicitly exploiting temporal continuity and active occlusion handling (Fig.~\ref{fig:network}). The model comprises two main parts: (i) a feature/correlation module and (ii) a recurrent fusion module that iteratively updates scene flow and disparity using the proposed Geometry-Motion Features (GMFs).

\vspace*{0.5em}
\noindent\textbf{1) Feature Encoding and Correlation Retrieval:}\label{Method-1-1}
We follow the RAFT architecture~\cite{teedRAFTRecurrentAllPairs2021} for feature extraction. A feature encoder $f_\theta$ produces 256-dimensional maps at 1/8 resolution, while a context encoder extracts reference frame semantics. 

We construct forward and backward 4D correlation pyramids $\{C^k\}_{k=1}^4$ between $I_t$ and $I_{t \pm 1}$ following~\cite{teedRAFTRecurrentAllPairs2021}. Correlation features $\text{Corr}^n$ are sampled via bilinear interpolation at projected positions $p' = K(\hat{d}_t^n K^{-1} p + s_t^n)$, where $\hat{d}_t^n$ and $s_t^n$ denote the depth and scene flow at iteration $n$, and $K$ is the camera intrinsic matrix.

\begin{figure*}[h]
	\centering
	\includegraphics{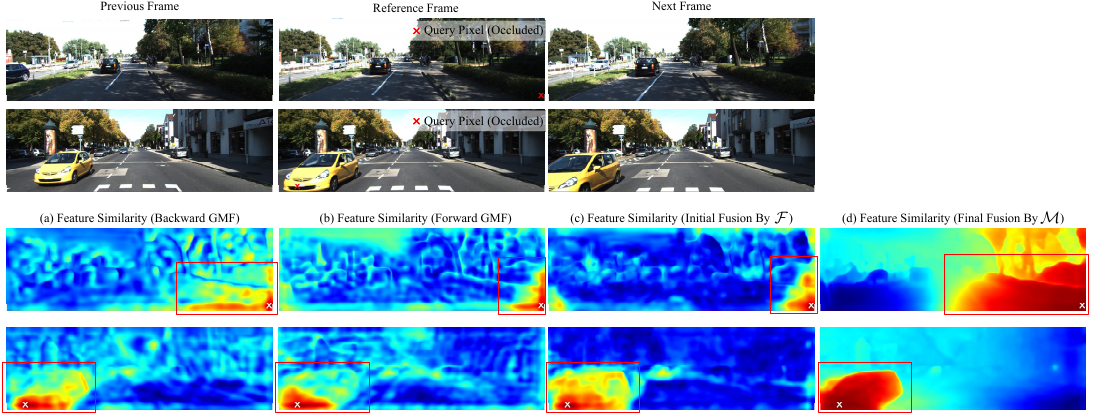}
	\caption{Visualization of GMF fusion. We compare the feature similarity maps of: (a) the backward GMF, (b) the forward GMF, (c) the initial fused GMF output by the fusion module $\mathcal{F}$, and (d) the final motion feature processed by the encoder $\mathcal{M}$ at an occluded query point (marked with X). Red regions indicate high feature similarity. While the forward feature is corrupted by occlusion, the backward feature remains reliable. The progression from (c) to (d) demonstrates how our framework effectively leverages backward cues to restore and refine feature consistency in the forward direction.}
	\label{fig:gmf_fusion}
\end{figure*}

\vspace*{0.5em}
\noindent\textbf{2) Recurrent Temporal Geometry-Motion Fusion:}

%Multi-frame scene flow estimation requires effective temporal integration of motion cues. 
We present a recurrent geometry-motion fusion framework that jointly estimates disparity and scene flow via temporal encoding, bidirectional fusion, and iterative refinement.

\noindent\textbf{Geometry-Motion Feature Fusion:} Unlike methods that rely on bidirectional correlation volumes or bidirectional flow, we adopt the Geometry-Motion Feature (GMF) for bidirectional interaction. As illustrated in Fig.~\ref{fig:update}, GMF is a unified representation extracted from GRU hidden states $\mathbf{h}^{\text{f}, n}$ and $\mathbf{h}^{\text{b}, n}$ via a projection head (three convolutional layers with ReLU):
\begin{equation}
	\hat{f}_{\text{GM}}^{\text{f}, n} = \text{Proj}(\mathbf{h}^{\text{f}, n}), \quad \hat{f}_{\text{GM}}^{\text{b}, n} = \text{Proj}(\mathbf{h}^{\text{b}, n}).
	\label{eq:gmf}
\end{equation}
GMF is initialized with learnable parameters at the first iteration and recursively updated in subsequent steps. The hidden state $\mathbf{h}$ is updated by a Gated Recurrent Unit (GRU) with current scene flow (motion) and disparity (geometry) as inputs, aggregating motion and geometry information over time. The forward and backward GMFs are bidirectionally fused in the motion feature encoder to guide estimation.

As shown in Fig.~\ref{fig:update}, the motion feature encoder contains standard components: the scene flow encoder, the optical flow encoder, the disparity encoder and the correlation encoder, each extracting uni-directional motion features for structural and dynamic cues. 
For temporal feature interaction, we design a lightweight temporal fusion module $\mathcal{F}$ (two convolutional layers with ReLU). As depicted in Fig.~\ref{fig:update}, it takes concatenated forward/backward GMFs as input:
\begin{equation}
	f_{\text{GM}}^{\text{f}, n} = \mathcal{F}\left([\hat{f}_{\text{GM}}^{\text{f}, n},\ -\hat{f}_{\text{GM}}^{\text{b}, n}]\right),
	\label{eq:gmf_fusion}
\end{equation}
where $[\,\cdot,\,\cdot\,]$ denotes channel-wise concatenation. We negate the backward GMF to align temporal directions ($t \to t+1$ and $t \to t-1$), reducing ambiguity and enforcing temporal consistency. The backward fused feature $f_{\text{GM}}^{\text{b}, n}$ is computed symmetrically.

We visualize GMF similarity maps in Fig.~\ref{fig:gmf_fusion} to verify the backward feature's complementarity. For an occluded query point, the forward GMF (Fig.~\ref{fig:gmf_fusion}(b)) produces weak cues due to occlusion, while the backward GMF (Fig.~\ref{fig:gmf_fusion}(a)) maintains high similarity. The fused feature (Fig.~\ref{fig:gmf_fusion}(c)) significantly expands high-similarity regions. 
The GMF design outperforms bidirectional correlation volumes/flow methods~\cite{hur2021self, mehl2023m} in temporal cue encoding, validated by ablation studies in Section~\ref{E-2}.

The fused feature $f_{\text{GM}}^{\text{f}, n}$ is concatenated with complementary features ($f_{\text{sf}}^{\text{f}, n}$, $f_{\text{of}}^{\text{f}, n}$, $f_{\text{d}}^{\text{f}, n}$, $f_{\text{c}}^{\text{f}, n}$) and processed by the motion encoder $\mathcal{M}$:
\begin{equation}
	\mathbf{M}^{\text{f}, n} = \mathcal{M}\left(f_{\text{sf}}^{\text{f}, n},\; f_{\text{of}}^{\text{f}, n},\; f_{\text{d}}^{\text{f}, n},\; f_{\text{c}}^{\text{f}, n},\;f_{\text{GM}}^{\text{f}, n}\right).
\end{equation}

As shown in Fig.~\ref{fig:gmf_fusion}(d), $\mathcal{M}$ outputs refined motion features with clearer structure than the raw forward GMF (Fig.~\ref{fig:gmf_fusion}(b)). The backward feature $\mathbf{M}^{\text{b}, n}$ uses the same fusion strategy.

\noindent\textbf{Recurrent Update Module:}
As presented in Fig.~\ref{fig:update}, this module is built on a GRU, which iteratively refines GMF, scene flow and disparity by updating hidden states with position-enhanced motion features, aggregated motion features and context features. The GRU captures temporal dependencies and corrects estimation errors:
\begin{equation}
	\begin{cases}
		\mathbf{h}^{\text{f}, n+1} = \mathcal{G}\left([\mathbf{M}_{\text{p}}^{\text{f}, n},\ \mathbf{M}^{\text{f}, n},\ g],\ \mathbf{h}^{\text{f}, n}\right), \\
		\mathbf{h}^{\text{b}, n+1} = \mathcal{G}\left([\mathbf{M}_{\text{p}}^{\text{b}, n},\ \mathbf{M}^{\text{b}, n},\ g],\ \mathbf{h}^{\text{b}, n}\right),
	\end{cases}
\end{equation}
where $\mathbf{h}^{\text{f}/\text{b}, n}$ are forward/backward hidden states, $g$ is the reference context feature (initializing $\mathbf{h}^{\text{f}/\text{b}, 0}$). Position-enhanced features $\mathbf{M}_{\text{p}}^{\text{f}/\text{b}, n}$ are detailed in Section~\ref{Method-2-1}.

The updated hidden state decodes GMF (Eq.~\ref{eq:gmf}) and predicts scene flow/disparity residuals via three convolutional layers. Residuals are added to current estimates for refinement:
\begin{equation}
	s_{t}^{\text{f}, n+1} = \Delta s_{t}^{\text{f}, n} + s_{t}^{\text{f}, n}, \quad d_{t}^{\text{f}, n+1} = \Delta d_{t}^{\text{f}, n} + d_{t}^{\text{f}, n},
\end{equation}
\begin{equation}
	s_{t}^{\text{b}, n+1} = \Delta s_{t}^{\text{b}, n} + s_{t}^{\text{b}, n}, \quad d_{t}^{\text{b}, n+1} = \Delta d_{t}^{\text{b}, n} + d_{t}^{\text{b}, n}.
\end{equation}

Initial scene flow estimates $s_{t}^{\text{f}/\text{b},0}$ are zero; initial disparity $d_{t}^{0}$ is generated from the reference image feature map. The final disparity is averaged: $d_t = (d_t^\text{f} + d_t^\text{b})/2$.

A learnable hidden-feature-based upsampler resolves the resolution gap, upsampling predictions from $1/8$ resolution to the original input resolution.

\subsection{Position-Enhanced Motion Feature Aggregation}\label{Method-2-1}

Inspired by the Global Motion Aggregation (GMA) module~\cite{jiang2021learning}, we extend position-only attention to multi-frame scene flow estimation for robust GMF propagation under occlusion. The attention weights are computed solely from relative spatial positions, independent of appearance.

%As shown in Fig.~\ref{fig:positional_att}, 
The attention map is defined as:
\begin{equation}
	\mathbf{A} = \text{Softmax}\left( \frac{Q(g)\cdot P^\top}{\sqrt{D}} \right),
\end{equation}
where $\mathbf{A} \in \mathbb{R}^{S \times S}$, $Q(g) \in \mathbb{R}^{S \times D}$ is the linearly projected query feature from context feature, and $P \in \mathbb{R}^{S \times D}$ is the relative positional embedding matrix with $S = H \times W$. 
For spatial position pair $(i,j)$ and $(i',j')$, the embedding is:
\begin{equation}
	p_{(i,j),(i',j')} = p_h(i'-i) + p_w(j'-j),
\end{equation}
where $p_h(\cdot)$ and $p_w(\cdot)$ are learnable embeddings for vertical and horizontal offsets. The matrix $P$ encodes all spatial relations, injecting geometric priors into attention.

The forward position-enhanced motion feature is:
\begin{equation}
	\mathbf{M}_{\text{p}}^{\text{f}, n}
	= \mathbf{M}^{\text{f}, n}
	+ \alpha \cdot \mathbf{A} \cdot \sigma(\mathbf{M}^{\text{f}, n}),
\end{equation}
where $\mathbf{M}^{\text{f}, n}$ is the motion feature, $\sigma(\cdot)$ is a linear projection, and $\alpha$ is a learnable scalar (initialized to zero). The backward feature $\mathbf{M}_{\text{p}}^{\text{b}, n}$ is computed analogously.

By modeling long-range spatial dependencies via positional priors, this module enables reliable feature propagation across spatially correlated regions, improving robustness in occluded and ambiguous areas.

\subsection{Self-supervised Loss}\label{Method-2-2}

We adopt the self-supervised loss functions proposed by Hur and Roth~\cite{hur2020self, hur2021self} and additionally introduce an occlusion regularization term to improve learning in occluded regions.

During training, given stereo pairs, the network takes two temporal triplets: \( \{I_{t-1}, I_{t}, I_{t+1}\} \) and \( \{I_{t}, I_{t+1}, I_{t+2}\} \), forming a 4-frame sequence.
We apply self-supervised loss on adjacent frames \( I_{t} \) and \( I_{t+1} \), using disparities \( (d_{t}, d_{t+1}) \) and scene flows \( (s_{t}^{\text{f}}, s_{t+1}^{\text{b}}) \).
Notably, during inference, the model uses only three consecutive monocular frames.

\noindent\textbf{Standard Losses:} 
Following prior work~\cite{hur2020self, hur2021self, godard2019digging}, we adopt standard self-supervised losses for disparity and scene flow, denoted as $L_d$ and $L_{\text{sf}}$, respectively, including photometric consistency, geometric/3D consistency, and edge-aware smoothness. All terms are computed over non-occluded regions using occlusion masks derived from disparity and scene flow, denoted as $O_{t}^{\text{d}}$ and $O_{t}^{\text{sf}}$, respectively.

\begin{figure}[t]
	\centering
	\includegraphics{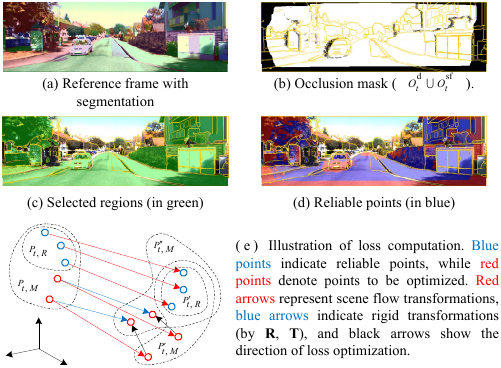}
	\caption{Illustration of the occlusion regularization process.}
	\label{fig:occ_reg}
%	\vspace{-1em}
\end{figure}

\noindent\textbf{Occlusion Regularization:}
Standard losses ignore occluded regions, limiting the network’s ability to exploit the GMF and learn accurate motion under occlusion. To address this, we introduce an occlusion regularization term that propagates reliable motion from visible regions.

As shown in Fig.~\ref{fig:occ_reg}, we segment the reference frame using SAM~\cite{kirillov2023segment}, assigning each pixel to a region mask corresponding to a common object (Fig.~\ref{fig:occ_reg}(a)). 
We then identify region masks containing occluded points (Fig.~\ref{fig:occ_reg}(c)) using the disparity mask \( O_{t}^{\text{d}} \) and scene flow mask \( O_{t}^{\text{sf}} \) (Fig.~\ref{fig:occ_reg}(b)). 
For each region \( M \), we define a reliable point mask \( O_R \) as:
\begin{equation}
	\begin{cases}
		O_R = M \cdot (1 - O_{t}^{\text{d}}) \cdot (1 - O_{t}^{\text{sf}}) \cdot O_P, \\
		O_P = \|P_t^{'} - P_{t+1}\|_2 < \theta,
	\end{cases}
\end{equation}
where $P_t^{'}$ and $P_{t+1}$ denote the warped 3D point from the reference frame and its corresponding 3D point in the target frame~\cite{hur2020self, hur2021self}. Thus, reliable points lie within $M$, are non-occluded, and satisfy a 3D consistency constraint (error \( < \theta \)). 
To remove unreliable distant regions (e.g., sky), we discard points with depth $>75$m and regions whose average reliable depth exceeds $25$m.

Given reliable points $p_R$, their 3D coordinates are:
\begin{equation}
	P_{t, R} = \hat{d}_t(p_R) \cdot K^{-1} p_R.
\end{equation}

Their counterparts $P_{t,R}'$ in the target frame are obtained via scene flow warping~\cite{hur2020self, hur2021self}. A rigid transformation $(R, T)$ is estimated via SVD between $P_{t, R}$ and $P_{t, R}'$ and applied to all points in $M$:
\begin{equation}
	\begin{cases}
		P_{t, M}^{''} = R \cdot P_{t, M} + T, \\
		P_{t, M} = \hat{d}_t(p_M) \cdot K^{-1} p_M.
	\end{cases}
\end{equation}

Meanwhile, 3D coordinates predicted via scene flow are:
\begin{equation}
		P_{t, M}^{'} = P_{t, M} + s_t^{\text{f}}(p_M).
\end{equation}

Finally, as shown in Fig.~\ref{fig:occ_reg}(e), the occlusion regularization loss is defined as:
\begin{equation}
	L_{\text{occ}} = \left\| \Theta(P_{t, M}^{''}) - P_{t, M}^{'} \right\|_2,
\end{equation}
where $\Theta$ stops gradients through $(R, T)$.

\noindent\textbf{Final Loss:}  
\begin{equation}
	\label{eq:floss}
	L_{\text{Total}} = \lambda_{\text{occ}} L_{\text{occ}} + \sum_{i=1}^{N} \zeta^{N-i} \left( L_d^i + \lambda_{\text{sf}} L_{\text{sf}}^i \right),
\end{equation}
where $N$ is the number of iterations, $\zeta$ is a decay factor, and $\lambda_{\text{occ}}$, $\lambda_{\text{sf}}$ balance the loss terms~\cite{hur2020self, hur2021self}. $L_{\text{occ}}$ is applied only at the final iteration due to early-stage instability.

%% file: sec/4_experiments.tex
\section{Experiments}
\subsection{Implementation Details}
\noindent\textbf{Dataset:} 
Following prior work~\cite{hur2020self, bayramli2023raft}, we train on KITTI raw~\cite{geiger2013vision} using the split of~\cite{godard2017unsupervised} and the protocol of Self-Mono-SF~\cite{hur2020self}. 
To avoid overlap with the KITTI Scene Flow Training set~\cite{Menze_2015_CVPR}, 29 overlapping scenes are removed, leaving 32 scenes (25{,}801 sequences). 
We further retain sequences with at least four consecutive frames, resulting in 22{,}040 training samples. 
The model is validated on the KITTI Scene Flow Training set and evaluated on the official testing benchmark. 

Following~\cite{hur2020self}, we semi-supervisedly fine-tune our model on the KITTI Scene Flow training set after self-supervised pre-training on the KITTI split.
%Training uses 4-frame sequences, while inference requires only 3 frames (Sec.~\ref{Method-2-2}).

\noindent\textbf{Training:} 
The model is implemented in PyTorch and optimized with AdamW~\cite{loshchilov2018decoupled} ($\beta_1=0.9$, $\beta_2=0.99$, weight decay $1\mathrm{e}{-4}$). 
The learning rate is initialized at $1\mathrm{e}{-4}$ with cosine annealing, and gradient clipping (norm 1.0) is applied. 
We train for 200k iterations with batch size 4, without employing multi-stage training. 
The number of update iterations is $N=10$ with decay factor $\zeta=0.8$. 
We set \( \lambda_{\text{occ}} = 1 \) and \( \theta = 0.025 \), while \( \lambda_{\text{sf}} \) is dynamically adjusted. 
Occlusion regularization is activated after 50\% of training to ensure sufficient reliable points. 
To mitigate checkerboard artifacts, we follow~\cite{bayramli2023raft} and detach gradients in the upsampler.

We apply photometric (gamma, brightness, color jitter) and geometric (flip, scaling, cropping) augmentations~\cite{hur2020self}. 
All images are resized to $256 \times 832$.

\input{sec/benchmark_train}

\input{sec/benchmark_test}

\begin{figure*}[htbp]
	\centering
	\includegraphics{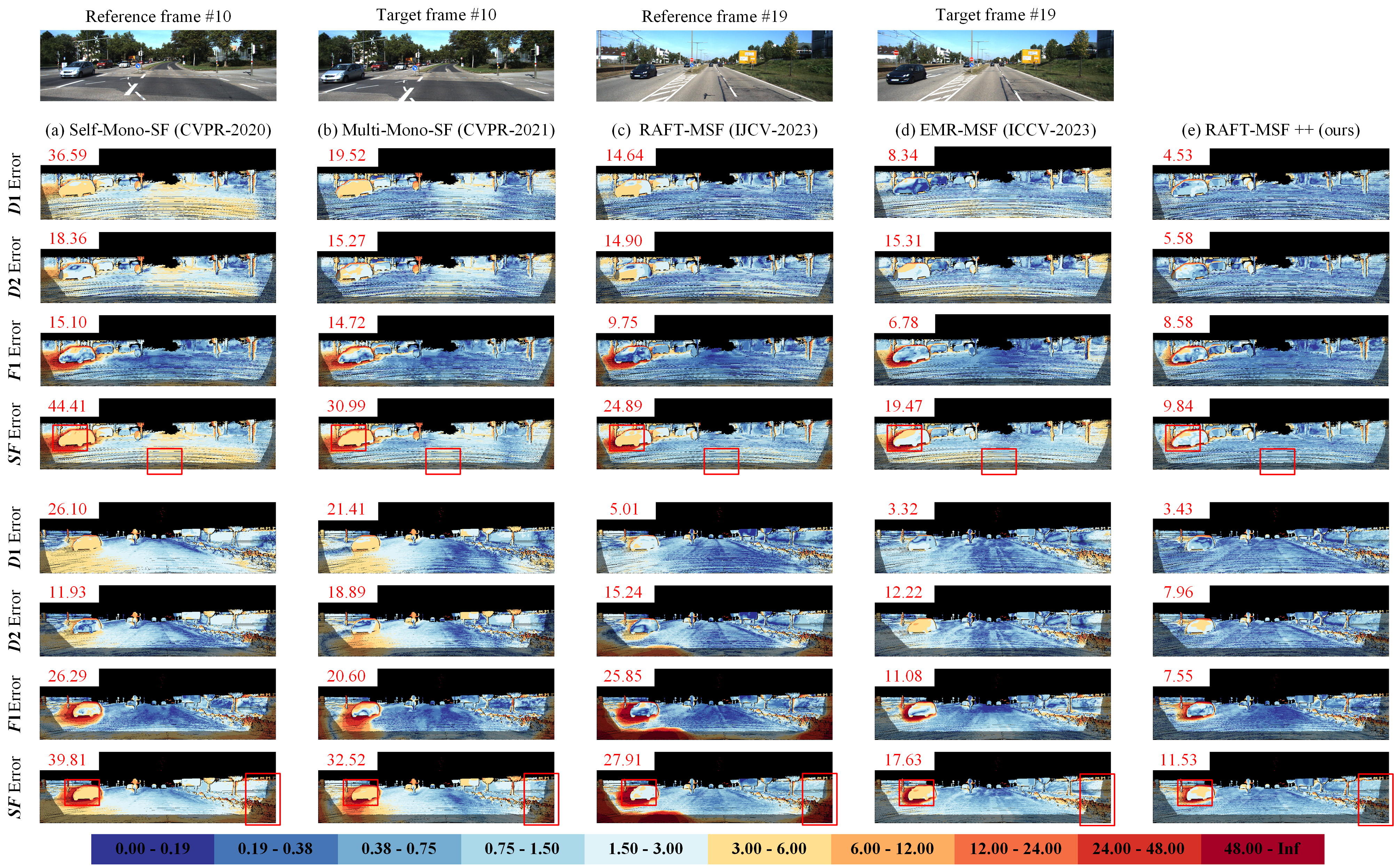}
	\caption{
		Qualitative results on the KITTI Scene Flow benchmark. We compare with Self-Mono-SF~\cite{hur2020self}, Multi-Mono-SF~\cite{hur2021self}, RAFT-MSF~\cite{bayramli2023raft}, and EMR-MSF~\cite{jiang2023emr}. Blue/red denote correct/incorrect predictions, and dark regions indicate occlusion. Outlier rates are overlaid on the corresponding error maps. Additional qualitative results are available on the official KITTI benchmark website.
	}
	\label{fig:qualitative}
%	\vspace*{-1em}
\end{figure*}

\noindent\textbf{Evaluation Metrics:} 
We follow the KITTI Scene Flow benchmark~\cite{Menze_2015_CVPR} and report outlier rates (\%). A pixel is considered an outlier if the error exceeds 3 pixels or 5\% of the ground truth. 
We report D1-all, D2-all, Fl-all, and the overall SF-all, computed on pixels valid across all components. 
Additionally, we report SF-noc and SF-occ for non-occluded and occluded regions. 
Unlike KITTI, we define occlusion as pixels invalid in any of the three components (D1, D2, or Fl), providing a stricter evaluation.

\subsection{Benchmark Testing} 

We compare our method with state-of-the-art monocular scene flow approaches on the KITTI Scene Flow training and testing benchmarks, with results summarized in Tabs.~\ref{tab:kitti_training} and~\ref{tab:kitti_testing}. To enable a structured comparison, we categorize existing methods into composite multi-network and unified single-network architectures. In addition, we report model parameters and inference runtime on a single RTX 3090 to provide a comprehensive evaluation of both accuracy and efficiency.

On the KITTI training set (Tab.~\ref{tab:kitti_training}), RAFT-MSF++ establishes a new state of the art within the unified single-network category, outperforming recent approaches such as CTSD~\cite{bayramli2024integrating} and Mamba-SF~\cite{chen2025mamba} across all metrics. While the composite EMR-MSF~\citep{jiang2023emr} achieves the lowest global errors (e.g., D1-all and SF-all), it relies on a multi-stage design that combines several task-specific sub-networks, including a large SDFA-Net~\cite{zhou2022self} depth backbone and a RAFT-3D~\cite{teed2021raft} flow estimator, resulting in over 76.77M parameters.

In contrast, RAFT-MSF++ adopts a compact end-to-end architecture with a shared encoder for joint geometry and motion learning. Despite using only 8.19M parameters (approximately 10\% of EMR-MSF), our geometry-motion feature (GMF) and temporal fusion effectively resolve spatial ambiguities, enabling our model to outperform EMR-MSF in challenging occluded regions (SF-occ: 25.38\% vs. 28.47\%).

On the KITTI testing benchmark (Tab.~\ref{tab:kitti_testing}), RAFT-MSF++ achieves strong performance within the self-supervised single-network category, yielding a 30.99\% relative improvement in SF-all over RAFT-MSF (34.98\% $\rightarrow$ 24.14\%). Qualitative results in Fig.~\ref{fig:qualitative} further support this trend, showing fewer errors in occluded regions compared to the composite EMR-MSF. 

Under supervised fine-tuning, our model attains an SF-all of 14.53\%, significantly outperforming the fine-tuned baseline (22.50\%) and the multi-stage Mono-SF~\cite{brickwedde2019mono}. Notably, our method achieves superior accuracy without relying on the 20k pseudo-ground-truth depth labels required by Mono-SF, highlighting the effectiveness of our unified learning framework.

Beyond accuracy, our unified architecture is highly efficient, requiring only 0.20s for bidirectional estimation. Moreover, our multi-frame GMF is designed as a modular, plug-and-play component rather than a rigidly engineered system.

To further validate this extensibility, we integrate our GMF module into RAFT-3D~\cite{teed2021raft}, the motion backbone of EMR-MSF~\citep{jiang2023emr} (whose official implementation is unavailable). As shown in Tab.~\ref{tab:kitti_testing}, our multi-frame component reduces the SF-all error of RAFT-3D from 5.77\% to 4.75\%. 
This demonstrates that our method provides a general temporal fusion mechanism that can benefit complex multi-stage pipelines such as RAFT-3D and EMR-MSF.

\input{sec/abla00}

\subsection{Ablation Study}\label{E-2}

We conduct ablation studies on the KITTI split to evaluate RAFT-MSF++ components. For efficiency, training is limited to 100k iterations. RAFT-MSF~\cite{bayramli2023raft} serves as the baseline; all variants use identical settings unless noted, without occlusion regularization or position-enhanced feature aggregation. The default configuration is marked in gray.

\noindent\textbf{Main Contributions:} Tab.~\ref{tab0} summarizes the impact of our key components: multi-frame estimation, occlusion regularization (Occ. Reg.), and relative positional attention (Rel. Pos. Attn.). 
Compared to the two-frame baseline, their combination yields a 36.6\% relative improvement in SF-all, with particularly strong gains on SF-occ, indicating improved robustness under occlusion.

Interestingly, relative positional attention degrades performance in the two-frame setting (Row~1 vs.~Row~4), but becomes effective when combined with multi-frame estimation (Row~2 vs.~Row~6), reducing SF-occ from 43.13\% to 38.27\%. 
This suggests that directly applying GMA~\cite{jiang2021learning} to unsupervised scene flow is insufficient, and that richer temporal cues are necessary to stabilize attention learning. Further analysis is provided in the supplementary material.

\input{sec/abla01}
\input{sec/abla02}

%\begin{figure*}[h]
%	\centering
%	\includegraphics{pic/ablation_main}
%	\caption{Qualitative results from the ablation study illustrating the impact of our main contributions. 
%%	The top row shows the reference frame and the forward target frame. 
%	Columns (a)–(e) depict results from different model variants: (a) the full model, (b) with multi-frame input and relative positional attention, (c) with multi-frame input and occlusion regularization, (d) with multi-frame input only, and (e) the baseline without either component. Each row presents disparity maps and their error maps for the reference frame (D1/D1 Error), warped disparity maps and error maps for the target frame (D2/D2 Error), optical flow and its error maps (Fl/Fl Error), and scene flow error maps (SF Error). In the error maps, correct estimations are indicated in \colorbox{blue}{\textcolor{white}{blue}}, incorrect estimations in \colorbox{red}{\textcolor{white}{red}}, and occluded pixels are shown as dark regions. Outlier rates are overlaid on the corresponding error maps.}
%	
%	
%	
%	
%	\label{fig:ablation_main}
%\end{figure*}

\noindent\textbf{Multi-Frame Extension:} 
We conduct an ablation study on extending the two-frame baseline to a multi-frame setting. Following prior work~\cite{hur2021self, mehl2023m, shi2023videoflow, dong2024memflow}, we adapt RAFT-MSF\footnote{\url{https://github.com/Bayrambai/raft-msf}}~\cite{bayramli2023raft} into several variants. All variants share the same backbone and differ only in multi-frame feature construction and fusion. 

Specifically, we implement: (i) a Multi-Mono-SF-like variant\footnote{\url{https://github.com/visinf/multi-mono-sf}}~\cite{hur2021self} with bidirectional correlation volumes (without LSTM), (ii) an M-FUSE-like variant\footnote{\url{https://github.com/cv-stuttgart/M-FUSE}}~\cite{mehl2023m} with staged training, and (iii) VideoFlow-like\footnote{\url{https://github.com/XiaoyuShi97/VideoFlow}}~\cite{shi2023videoflow} and MemFlow-like\footnote{\url{https://github.com/DQiaole/MemFlow}}~\cite{dong2024memflow} variants. We also include RAFT-MSF++ with bidirectional correlation volumes.

\input{sec/abla03}
\input{sec/abla04}

\begin{table}[tb]
	\caption{Performance with different iteration numbers.}
	\label{tab:iter_ablation}
	\centering
	\fontsize{9pt}{11pt}\selectfont
	\begin{tabular}{lccccc}
		\toprule
		Iter. Num. & D1-all & D2-all & Fl-all & SF-all  & Runtime (s) \\
		\midrule
		1  & 40.76 & 50.67 & 34.91 & 64.07 & 0.05 \\
		2  & 31.22 & 36.05 & 25.17 & 47.98 & 0.07 \\
		4  & 18.01 & 20.89 & 16.23 & 28.76 & 0.09 \\
		6  & 14.69 & 18.21 & 12.83 & 23.90 & 0.13 \\
		8  & \textbf{13.90} & 17.35 & \textbf{11.93} & 22.54 & 0.16 \\
		\rowcolor{gray!20}
		10 & \underline{13.95} &  \underline{16.86} & \underline{12.09} & \underline{21.96} & 0.20 \\
		12 & 14.38 & \textbf{16.70} & 12.73 & \textbf{21.86} & 0.23 \\
		\bottomrule
	\end{tabular}
	\vspace{-1em}
\end{table}

As shown in Tab.~\ref{tab1}, incorporating bidirectional volumes into the Multi-Mono-SF-like model significantly improves performance, which contrasts with the original Multi-Mono-SF~\cite{hur2021self} paper; further analysis is provided in the supplementary material. The M-FUSE-like, VideoFlow-like, and MemFlow-like variants also yield consistent improvements, indicating the benefit of richer temporal context modeling.

However, these methods provide limited gains in D1-all and D2-all, with most improvements coming from Fl-all. This suggests that they primarily enhance motion (2D flow) estimation, while offering weaker improvements for geometric reasoning across frames. 
In contrast, RAFT-MSF++ improves all metrics, achieving a 27.29\% relative reduction in SF-all, demonstrating the effectiveness of our Geometry-Motion Feature (GMF) for joint modeling of geometry and motion.

Notably, adding bidirectional correlation volumes to RAFT-MSF++ degrades most metrics except Fl-all. We attribute this to feature redundancy, which interferes with GMF learning and weakens geometric feature propagation, further indicating that correlation volumes mainly contribute motion cues rather than geometric consistency.

\noindent\textbf{Geometry-Motion Feature:} 
We analyze both feature extraction and fusion strategies for geometry–motion features.

For feature extraction, we compare (i) splitting features from the motion encoder output and (ii) decoding from GRU hidden states. As shown in Tab.~\ref{tab2}, both outperform the baseline, while GRU-based decoding achieves the best results. This indicates that recurrent hidden states provide a more expressive geometry–motion representation, benefiting from temporal aggregation across iterations, in contrast to the MotionEncoder, which relies on local correlations.

For temporal fusion, we evaluate addition, gated attention, and concatenation (Tab.~\ref{tab3}). Concatenation consistently achieves the best performance across all metrics.

\noindent\textbf{Number of Input Frames:}
Our method uses three-frame fusion, while prior work suggests that using more frames may further improve performance~\cite{hur2021self, shi2023videoflow, sun2025m2flow}. To investigate this, we extend the baseline to four-frame variants following representative designs. Specifically, we construct three variants: a VideoFlow-style extension~\cite{shi2023videoflow}, a Multi-Mono-SF-based variant~\cite{hur2021self} with LSTM integration, and an M2Flow-based variant\footnote{\url{https://github.com/sunzunyi/M2FLOW}}~\cite{sun2025m2flow} with Motion Information Propagation (MIP), both designed to propagate Geometry–Motion Features (GMF) across frames.

As shown in Tab.~\ref{tab4}, four-frame variants generally underperform three-frame ones. We attribute this to feature misalignment: three-frame fusion aligns forward and backward features to a central frame, ensuring stable correspondences, whereas multi-frame fusion requires warping features to neighboring frames via optical flow~\cite{hur2021self, shi2023videoflow, sun2025m2flow}, which is reliable for motion but less so for geometry.

Since scene flow jointly models geometry and motion, depth features are not temporally consistent under camera and object motion~\cite{liang2025zeroSF}. As a result, warping introduces noise into geometry features and degrades performance.

In summary, geometry and motion are tightly coupled, and extending beyond three frames amplifies misalignment, particularly in geometry estimation. This motivates our choice of three-frame fusion.

Interestingly, this observation differs from~\cite{hur2021self}, which reports gains with longer sequences (up to five frames). We provide further analysis in the supplementary material.

%\noindent\textbf{Iteration Number:}
%We evaluate the impact of the number of update iterations on accuracy and runtime (Tab.~\ref{tab:iter_ablation}). 
%As the number of iterations increases from 1 to 12, the scene flow error decreases rapidly in early iterations and saturates around 10 iterations. 
%We therefore select 10 iterations as a good trade-off between accuracy and efficiency. In addition, our model requires 2.36 GB of GPU memory for inference at a resolution of $256 \times 832$.

\noindent\textbf{Iteration Number:}
We evaluate the effect of iteration number on accuracy and runtime (Tab.~\ref{tab:iter_ablation}). The error decreases rapidly in early iterations and saturates around 10 iterations, which we select as the default setting for a good accuracy-efficiency trade-off. In addition, our model requires 2.36 GB GPU memory for inference at $256 \times 832$ resolution.

%% file: sec/benchmark_train.tex
\begin{table*}[t]
	\caption{
		Quantitative evaluation of the scene flow on the KITTI Scene Flow Training set. Metrics are evaluated over all pixels (\textit{all}), non-occluded regions (\textit{noc}), and occluded regions (\textit{occ}).  Missing entries (-) indicate results not reported. All values are reported as lower-the-better.
	}
	\label{tab:kitti_training}
	\centering
	\fontsize{9pt}{11pt}\selectfont
	\setlength{\tabcolsep}{0pt} % 关闭自动间距，完全手动控制
	\begin{tabular}{
			% 手动设置每一列宽度，修改 cm 数值即可！
			p{3.5cm}    % Method 列（最宽）
			>{\centering\arraybackslash}p{1.6cm} % D1-all
			>{\centering\arraybackslash}p{1.6cm} % D2-all
			>{\centering\arraybackslash}p{1.6cm} % Fl-all
			>{\centering\arraybackslash}p{1.6cm} % SF-all
			>{\centering\arraybackslash}p{1.6cm} % SF-noc
			>{\centering\arraybackslash}p{1.6cm} % SF-occ
			>{\centering\arraybackslash}p{1.6cm} % Param.
			>{\centering\arraybackslash}p{1.6cm} % Runtime
		}
		\toprule[0.8pt]
		\quad Method & D1-all & D2-all & Fl-all & SF-all & SF-noc & SF-occ & Param. & Runtime\\
		\midrule[0.5pt]
		
		\multicolumn{9}{l}{\textbf{\textit{\quad Composite Multi-Network Architectures}}} \\
		\midrule
		\quad DF-Net~\citep{zou2018df} & 46.50 & 61.54 & 27.47 & 73.30 & $-$ & $-$ & 98.40M & $-$\\
		\quad GeoNet~\citep{yin2018geonet} & 49.54 & 58.17 & 37.83 & 71.32 & $-$ & $-$ & 118.50M & 0.02s \\
		\quad EPC++~\citep{luo2019every} & 23.84 & 60.32 & 19.64 & $>60.32$ & $-$ & $-$ & 39.58M & 0.02s \\
		\quad EMR-MSF~\citep{jiang2023emr} & \textbf{8.37} & \textbf{12.86} & \textbf{11.58} & \textbf{18.11} & \textbf{16.48} & \underline{28.47} & $>76.77$M & 0.25s  \\
		\midrule
		
		\multicolumn{9}{l}{\textbf{\textit{\quad Unified Single-Network Architectures}}}\\
		\midrule
		\quad Self-Mono-SF~\citep{hur2020self} & 31.25 & 34.86 & 23.49 & 47.05 & 41.25 & 73.41 & 5.76M & 0.03s \\
		\quad Multi-Mono-SF~\citep{hur2021self} & 27.33 & 30.44 & 18.92 & 39.82 & 34.63 & 62.65 & 7.54M & 0.21s \\
		\quad RAFT-MSF~\citep{bayramli2023raft} & 18.34 & 23.65 & 17.51 & 30.97 & 24.44 & 63.19 & 5.74M & 0.13s \\
		\quad GloFP-MSF~\citep{xiang2024glofp} & 22.43 & 23.67 & 17.72 &  31.32  & $-$ & $-$ & 10.21M & $-$\\
		\quad CTSD~\citep{bayramli2024integrating} & 16.77 & 21.74 & 15.89 &  28.50 & $-$ & $-$ & $>5.74$M & $-$\\
		\quad Mamba-SF~\citep{chen2025mamba} & 15.24 & 19.36 & 13.87 & 25.14 & $-$ & $-$ & $>5.74$M & $-$ \\
		\quad RAFT-MSF++ (ours) & \underline{13.95} & \underline{16.86} & \underline{12.09} & \underline{21.96} & \underline{21.11} & \textbf{25.38}  & 8.19M & 0.20s \\
		\bottomrule[0.8pt]
	\end{tabular}
	\\[2pt]
\end{table*}

%% file: sec/benchmark_test.tex
\begin{table*}[t]
	\caption{Quantitative evaluation of the scene flow on the KITTI Scene Flow Testing set. Metrics are evaluated over all pixels (\textit{all}).  Missing entries (-) indicate results not reported. All values are reported as lower-the-better.}
	\label{tab:kitti_testing}
	\centering
	\fontsize{9pt}{11pt}\selectfont
	\setlength{\tabcolsep}{0pt}
	\begin{tabular}{
			p{3.5cm}    % Method
			>{\centering\arraybackslash}p{1.8cm} % D1-all
			>{\centering\arraybackslash}p{1.8cm} % D2-all
			>{\centering\arraybackslash}p{1.8cm} % Fl-all
			>{\centering\arraybackslash}p{1.8cm} % SF-all
			>{\centering\arraybackslash}p{1.8cm} % Param.
			>{\centering\arraybackslash}p{1.8cm} % Runtime
		}
		\toprule[0.8pt]
		\quad Method & D1-all & D2-all & Fl-all & SF-all & Param. & Runtime\\
		\midrule[0.5pt]
		
		\multicolumn{7}{l}{\textbf{\textit{\quad Supervised Fine-Tuning}}} \\
		\midrule
		\quad Mono-SF~\citep{brickwedde2019mono} & 16.32 & 19.59 & 12.77 & 23.08 & $-$ & 41s \\
		\quad Self-Mono-SF-ft~\citep{hur2020self} & 22.16 & 25.24 & 15.91 & 33.88 & 5.76M & 0.03s \\
		\quad Multi-Mono-SF-ft~\citep{hur2021self} & 22.71 & 26.51 & 13.37 & 33.09 & 7.54M & 0.21s \\
		\quad RAFT-MSF-ft~\citep{bayramli2023raft} & \underline{15.95} & \underline{18.77} & \underline{8.80} & \underline{22.50} & 5.74M & 0.13s \\
		\quad RAFT-MSF++ -ft (ours) & \textbf{9.98} & \textbf{12.03} & \textbf{6.19} & \textbf{14.53} & 8.19M & 0.20s \\
		\midrule
		
		\multicolumn{7}{l}{\textbf{\textit{\quad Self-Supervised: Composite Multi-Network Architectures}}} \\
		\midrule
		\quad EMR-MSF~\citep{jiang2023emr} & \textbf{9.70} & \textbf{14.51} & \textbf{11.93} & \textbf{19.74} & $>76.77$M & 0.25s  \\
		\midrule
		
		\multicolumn{7}{l}{\textbf{\textit{\quad Self-Supervised: Unified Single-Network Architectures}}} \\
		\midrule
		\quad Self-Mono-SF~\citep{hur2020self} & 34.02 & 36.34 & 23.54 & 49.54 & 5.76M & 0.03s \\
		\quad Multi-Mono-SF~\citep{hur2021self} & 30.78 & 34.41 & 19.54 & 44.04 & 7.54M & 0.21s \\
		\quad RAFT-MSF~\citep{bayramli2023raft} & 21.21 & 27.51 & 18.37 & 34.98 & 5.74M & 0.13s \\
		\quad CTSD~\citep{bayramli2024integrating} & 20.69 & 27.36 & 17.32 &  33.92 & $>5.74$M & $-$\\
		\quad RAFT-MSF++ (ours) & \underline{16.11} & \underline{18.81} & \underline{12.19} & \underline{24.14} & 8.19M & 0.20s \\
		\midrule
		
		\multicolumn{7}{l}{\textbf{\textit{\quad Supervised Stereo Baseline (Plug-and-play validation)}}} \\
		\midrule
		\quad RAFT-3D~\citep{teed2021raft} & 1.81 & 3.67 & 4.29 & 5.77 & 44.50M & 0.25s \\
		\quad RAFT-3D-MF (ours) & 1.81 & \textbf{2.97} & \textbf{3.55} & \textbf{4.75} & 46.55M & 0.26s \\
		\bottomrule[0.8pt]
	\end{tabular}
\end{table*}

%% file: sec/abla00.tex
\begin{table*}[h]  %  [htbp]
		\caption{
			Ablation study on the effectiveness of our main components. 
			Metrics are evaluated over all pixels (\textit{all}), non-occluded regions (\textit{noc}), and occluded regions (\textit{occ}). 
			All values are reported as lower-the-better.
	}
	\label{tab0}%
	\centering
	\fontsize{9pt}{11pt}\selectfont
	%\vspace*{-1em}
	%	\resizebox{0.8\linewidth}{!}{
    \begin{tblr}{
			colspec={ccccccccc},
			stretch=0.5, 
			colsep=8.4pt,
			row{8} = {bg=gray!20},
			hline{1,Z} = {0.8pt},
			hline{2} = {0.5pt},
%			row{1} = {font=\bfseries}
		}
		Multi-Frame & Occ. Reg. & Rel. Pos. Attn. & D1-all & D2-all & Fl-all & SF-all & SF-noc & SF-occ \\
		\SetCell[r=1, c=3]{l}{\textit{(RAFT-MSF baseline)}} & & & 26.72 & 30.56 & 17.47 & 38.14 & 33.14 & 62.92 \\
		 \checkmark & & & 17.56 & 20.88 & 14.73 & 27.73 & 24.61 & 43.13 \\
		 & \checkmark & & 25.01 & 26.71 & 19.06 & 34.59 & 31.49 & 48.58 \\
		 &  & \checkmark & 46.80 & 52.39 & 17.19 & 57.63 & 51.36 & 90.50 \\
%		 & \checkmark & \checkmark & 35.93 & 36.10 & 18.30 & 43.38 & 37.99 & 69.93 \\
		\checkmark & \checkmark & & 17.70 & 20.07 & \textbf{12.87} & 25.42 & 24.07 & 31.53 \\
		\checkmark&  & \checkmark & \textbf{15.68} & 20.01 & 14.31 & 26.42 & 23.99 & 38.27 \\
		\checkmark & \checkmark & \checkmark & \underline{15.88} & \textbf{19.19} & \underline{12.88} & \textbf{24.17} & \textbf{23.37} & \textbf{27.56} \\
	\end{tblr}
		%	}
	%	\vspace*{-0.5em}
\end{table*}

%% file: sec/abla01.tex
\begin{table*}[t]  %  [htbp]
	\caption{Ablation study of multi-frame extension.}
	\label{tab1}%
	\centering
	\fontsize{9pt}{11pt}\selectfont
	%\vspace*{-1em}
	%	\resizebox{0.8\linewidth}{!}{
		\begin{tblr}{
				colspec={clcccc},
				stretch=0.5, 
				colsep=12.4pt,
				row{8} = {bg=gray!20},
				hline{1,Z} = {0.8pt},
				hline{2} = {0.5pt},
				%			row{1} = {font=\bfseries}
			}
			Frames & Method & D1-all & D2-all & Fl-all & SF-all \\
			2 & \textit{RAFT-MSF baseline} & 26.72 & 30.56 & 17.47 & 38.14 \\
			3 & Multi-Mono-SF \citep{hur2021self} & 24.57 & 28.97 & 16.46 & 35.30 \\
			3 & M-FUSE \citep{mehl2023m} & 22.52 & 28.22 & 21.51 & 38.13 \\
			3 & VideoFlow-TOF \citep{shi2023videoflow} & 23.87 & 26.36 & 15.91 & 33.01 \\
			3 & MemFlow \citep{dong2024memflow} & 25.43 & 32.07 & 17.18 & 37.87 \\
			3 & RAFT-MSF++ with Bid-correlation volumes & 20.35 & 23.49 & \textbf{14.41} & 29.44 \\
			3 & RAFT-MSF++(ours) & \textbf{17.56} & \textbf{20.88} & 14.73 & \textbf{27.73} \\
		\end{tblr}
		%	}
	
	%	\vspace*{-0.5em}
	\vspace*{-1em}
\end{table*}

%% file: sec/abla02.tex
%\vspace*{-1.2em}
\begin{table}[tb]
	\caption{Ablation study of geometry–motion feature extraction strategies.}
	\label{tab2}%
	\centering
		\fontsize{9pt}{11pt}\selectfont
	\begin{tabular}{lcccc}
		
		\toprule
		Extraction Strategy & D1-all & D2-all & Fl-all & SF-all \\
		\midrule
		None \textit{(baseline)}& 26.72 & 30.56 & 17.47 & 38.14 \\
		MotionEncoder split       & 23.10 & 26.49 & 15.90 & 32.66 \\
		\rowcolor{gray!20}
		GRU state decoding        & \textbf{17.56} & \textbf{20.88} & \textbf{14.73} & \textbf{27.73} \\
		\bottomrule
	\end{tabular}
	
\end{table}
%\vspace*{-2.0em}

%% file: sec/abla03.tex
%\vspace*{-1.2em}
\begin{table}[tb]
	\caption{Ablation study on geometry–motion feature fusion strategies.}
	\label{tab3}%
	\centering
	\fontsize{9pt}{11pt}\selectfont
%	\begin{tabular}{p{2.2cm}cccc}
%	\begin{tabular}{@{}lcccc@{}}
	\begin{tabular}{lcccc}
		\toprule
		Fusion Strategy & D1-all & D2-all & Fl-all & SF-all \\
		\midrule
		None \textit{(baseline)}& 26.72 & 30.56 & 17.47 & 38.14 \\
		Addition & 18.54 & 22.86 & 15.14 & 29.88 \\
		Gated & 20.73 & 23.22 & 14.25 & 29.05 \\
		\rowcolor{gray!20}
		Concatenation & \textbf{17.56} & \textbf{20.88} & \textbf{14.73} & \textbf{27.73} \\
		\bottomrule
	\end{tabular}
\end{table}

%\vspace*{-2.0em}

%% file: sec/abla04.tex
%\vspace*{-1.2em}
\begin{table*}[h]  %  [htbp]
	\caption{Ablation study on 4-frame input.}
	\label{tab4}%
	\centering
	\fontsize{9pt}{11pt}\selectfont
	%\vspace*{-1em}
	%	\resizebox{0.8\linewidth}{!}{
		\begin{tblr}{
%				colspec={X[c] l X[c] X[c] X[c] X[c]},
				colspec={clcccc},
				stretch=0.5, 
				colsep=12.4pt,
				row{6} = {bg=gray!20},
				hline{1,Z} = {0.8pt},
				hline{2,4} = {0.5pt}, 
				%			row{1} = {font=\bfseries}
			}
			Frames & Method & D1-all & D2-all & Fl-all & SF-all \\
			4 & VideoFlow-MOF \citep{shi2023videoflow} & 36.95 & 40.31 & 58.66 & 69.06 \\
			3 & VideoFlow-TOF \citep{shi2023videoflow} & \textbf{23.87} & \textbf{26.36} & \textbf{15.91} & \textbf{33.01} \\
			4 & RAFT-MSF++ with LSTM & 19.51 & 27.07 & 16.44 & 33.35 \\
			4 & RAFT-MSF++ with MIP & 19.55 & 27.65 & 21.99 & 36.43 \\
			3 & RAFT-MSF++ (ours) & \textbf{17.56} & \textbf{20.88} & \textbf{14.73} & \textbf{27.73} \\
		\end{tblr}
		%	}
	
		\vspace*{-1em}
\end{table*}

%% file: appendix.tex
%
%\setcounter{figure}{6}    % 图片编号重置为 1
%\setcounter{table}{8}     % 表格编号重置为 1

% 注意：如果使用 IEEEtran、 NeurIPS 等模板，需确保 \appendices 环境正确加载（多数模板支持）
\appendices  % 开启多附录模式（自动编号 A、B、C...），比单个 \appendix 更规范

% Empirical Analysis: 

\section{Multi-frame Context as a Prerequisite for Relative Positional Attention}
\label{sec:supp_attention}

To validate the mechanism of the proposed relative positional attention and explain the performance divergence between two-frame and multi-frame settings, we provide a qualitative visualization of the internal attention maps and feature similarity.

\begin{figure*}[htbp]
	\centering
	% 请替换为你生成的对比图路径
	\includegraphics[width=0.95\linewidth]{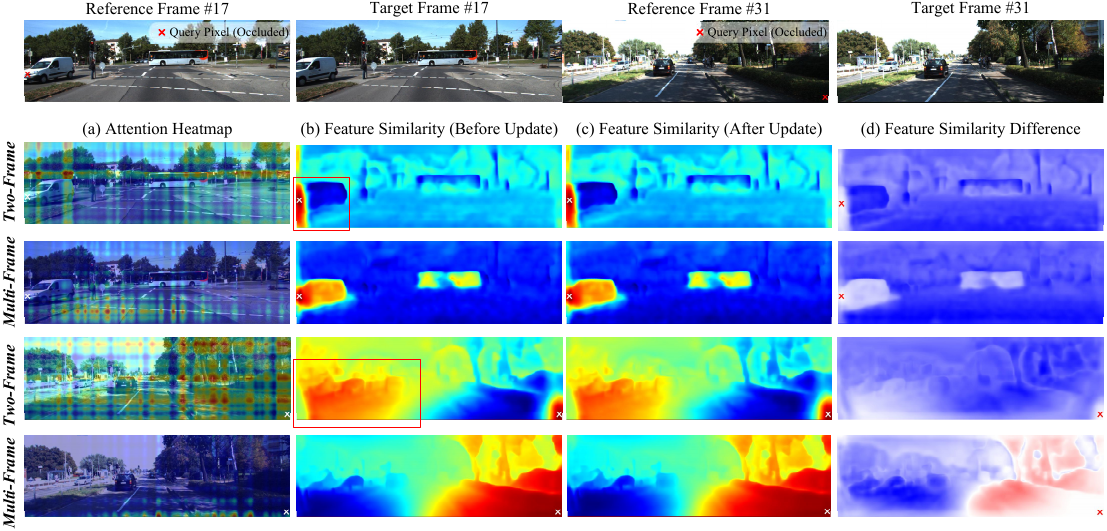}
	\caption{
	Visualization of Attention Maps and Feature Similarity. Top row: Input frames; Second and third rows: Results for Frame \#17; Fourth and fifth rows: Results for Frame \#31. 
	The query pixel (marked with X) is located in an occluded region. (a) Comparison of attention maps: The two-frame model shows grid-like noise, while the multi-frame model focuses attention on a horizontal band aligned with the query pixel, demonstrating that the model learns to attend to the spatially adjacent, visible parts of the object. (b-d) Feature similarity analysis: 
	The difference map in (d) shows that the multi-frame aggregation selectively enhances the feature similarity (red regions) between the occluded query and its reliable neighbors. Conversely, the two-frame attention either suppresses relevant features or preserves irrelevant ones due to the noisy attention map.
	}

	\label{fig:supp_att_analysis}
\end{figure*}

%\noindent\textbf{Mechanism Analysis:} 
As shown in Fig.~\ref{fig:supp_att_analysis}, we select two query pixels from different occluded regions to analyze the behavior of the attention module. These regions correspond to two primary types of occlusion: object-induced occlusions and occlusions caused by camera ego-motion.

\begin{itemize}

	\item \textbf{Failure in Two-frame Setting (Attention Map):} In the two-frame baseline, the attention weights exhibit a grid-like, noisy pattern. This indicates that without sufficient temporal cues, the model fails to establish reliable feature consistency, instead overfitting to the inherent biases of the learned positional embeddings rather than capturing the actual scene geometry.
	
	\item \textbf{Success in Multi-frame Setting (Attention Map):} In contrast, under the multi-frame setting, the attention map focuses distinctly along a horizontal band aligned with the query pixel. This demonstrates that the model effectively leverages temporal context to capture the object's spatial structure, ensuring that the mechanism searches for correspondences within physically plausible regions.

	\item \textbf{Input Feature Quality:} We further analyze the quality of motion features prior to the attention update (Fig.~\ref{fig:supp_att_analysis}(b)). The multi-frame geometry-motion features (GMFs) exhibit significantly higher spatial consistency between the occluded query and its reliable neighbors (darker red). 
	For instance, in the vehicle region, the multi-frame setting (Row 3, Col 2) maintains consistent feature semantics across the object, accurately reflecting its rigid-body nature. In contrast, the two-frame baseline (Row 2, Col 2) loses the vehicle's semantic integrity. 
	Furthermore, the two-frame features often display spurious high correlations with incorrect, spatially distant regions (as evidenced in Row 4, Col 2), while failing to capture valid local context. 
	This confirms that the multi-frame context provides a superior feature representation serving as a robust basis for the attention mechanism.
	
	\item \textbf{Feature Propagation (Aggregation Effect):} Fig.~\ref{fig:supp_att_analysis}(d) visualizes the change in feature similarity after aggregation. The red areas indicate increased similarity, confirming that the attention mechanism effectively propagates motion information from the visible parts of the object to the occluded query pixel. In contrast, the two-frame attention mechanism either suppresses relevant features or retains irrelevant ones due to its noisy attention map.
%	, thereby "repairing" the corrupted features.
\end{itemize}

In conclusion, the multi-frame setting is a prerequisite for the effective use of relative positional attention in unsupervised scene flow estimation. 
The richer temporal context enables the attention module to learn meaningful spatial structures and to reliably propagate motion cues from visible, reliable regions to occluded areas.

\section{Ablation Study on Occlusion Regularization Strategy}
\label{sec:supp_occ_loss}

We analyze the effect of occlusion regularization at different iterative stages (Tab.~\ref{tab:occ_loss}).
Applying the loss at all steps leads to a minor performance drop (SF-all +0.40\%). 
This small gap verifies the module acts as a soft geometric constraint rather than a rigid regularizer, maintaining stable training even on coarse early predictions. 
However, to achieve the best possible accuracy and reduce computational overhead, we employ the ``Final only'' strategy.

\begin{table}[htbp]
	\caption{
		Ablation study on the timing of occlusion regularization loss application.
	}
	\label{tab:occ_loss}%
	\centering
	\fontsize{9pt}{11pt}\selectfont
	%	\begin{tabular}{p{2.2cm}cccc}
		%	\begin{tabular}{@{}lcccc@{}}
			\begin{tabular}{lcccc}
				\toprule
				Strategy & D1-all & D2-all & Fl-all & SF-all \\
				\midrule
				All iterations  & 15.97 & 19.72 & \textbf{12.76} & 24.57 \\
				\rowcolor{gray!20}
				Final only & \textbf{15.88} & \textbf{19.19} & 12.88 & \textbf{24.17} \\
				\bottomrule
			\end{tabular}
\end{table}
		
\section{Analysis of Temporal Fusion Effectiveness in Multi-Mono-SF}
\label{sec:supp_multimono_fusion}

	In our ablation studies, we observed results that contradict key claims made in the original Multi-Mono-SF~\citep{hur2021self} paper. Specifically, while Multi-Mono-SF reports that incorporating bidirectional cost volumes yields only marginal improvements and that five-frame fusion via LSTM leads to better performance than three-frame input, our experiments reveal the opposite: bidirectional cost volumes significantly improve accuracy, and extending fusion beyond three frames consistently degrades performance.
	
	To investigate this discrepancy, we conduct a controlled experiment on the official Multi-Mono-SF model. Specifically, we manually zeroed out the LSTM's hidden and cell states at each time step, effectively disabling its temporal memory and reducing the module to a stateless nonlinear transformation. We then evaluated the modified model using a three-frame input, while keeping all other components unchanged and loading the officially released checkpoint.
	
	As summarized in Tab.~\ref{tab5}, the resulting performance is nearly identical to the original five-frame results reported in the paper, despite the absence of any temporal aggregation beyond three frames.
	
	These findings suggest that the improvements reported in the original Multi-Mono-SF paper may not be attributed to multi-frame temporal modeling via LSTM, but rather to the architectural benefits of the convolutional operations embedded within the LSTM. In particular, these convolutions likely enhance the integration of bidirectional cost volumes within a three-frame temporal window, which appears to be sufficient for the model's performance gains.
	
	\input{sec/abla05}

\begin{figure*}[htbp]
	\centering
	\includegraphics[width=1.0\linewidth]{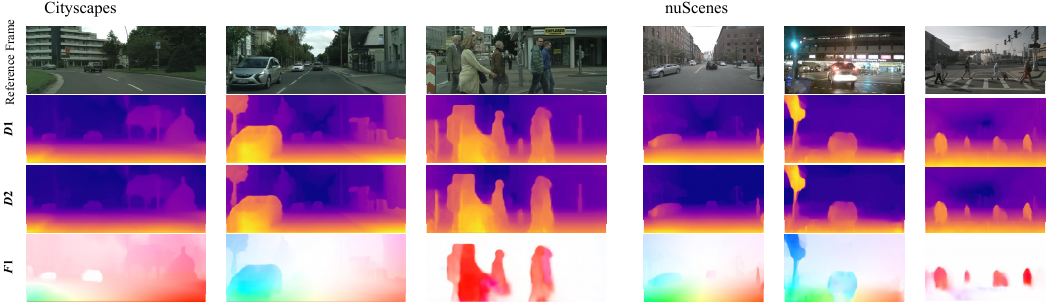}
	\caption{Zero-shot generalization results on Cityscapes~\cite{cordts2016cityscapes} and nuScenes~\cite{caesar2020nuscenes}. 
		The model is trained solely on KITTI and tested directly on these unseen datasets without fine-tuning. 
		Left: Cityscapes results demonstrate accurate motion estimation for pedestrians and dynamic urban traffic.
		Right: nuScenes results highlight robustness under challenging conditions, including nighttime scenes.
		This validates that our method learns robust physical priors applicable to diverse autonomous driving scenarios.
	}
	
	\label{fig:generalization}
\end{figure*}

\begin{figure}[t]
	\centering
	\includegraphics{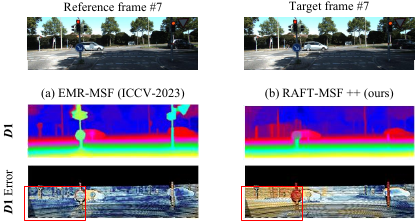}  %  [width=0.9\columnwidth]   [width=0.8\textwidth]
	\caption{
		A representative failure case on the KITTI Scene Flow Test benchmark caused by minimal ego-motion (e.g., vehicle stationary at a red light). Due to insufficient parallax between frames, stereo cues become weak, resulting in inaccurate disparity estimation. This error subsequently propagates to the scene flow prediction, particularly in low-motion or static regions.
	}
	
	\label{fig:limit}
\end{figure}

\begin{figure*}[htbp]
	\centering
	\includegraphics[width=1.0\linewidth]{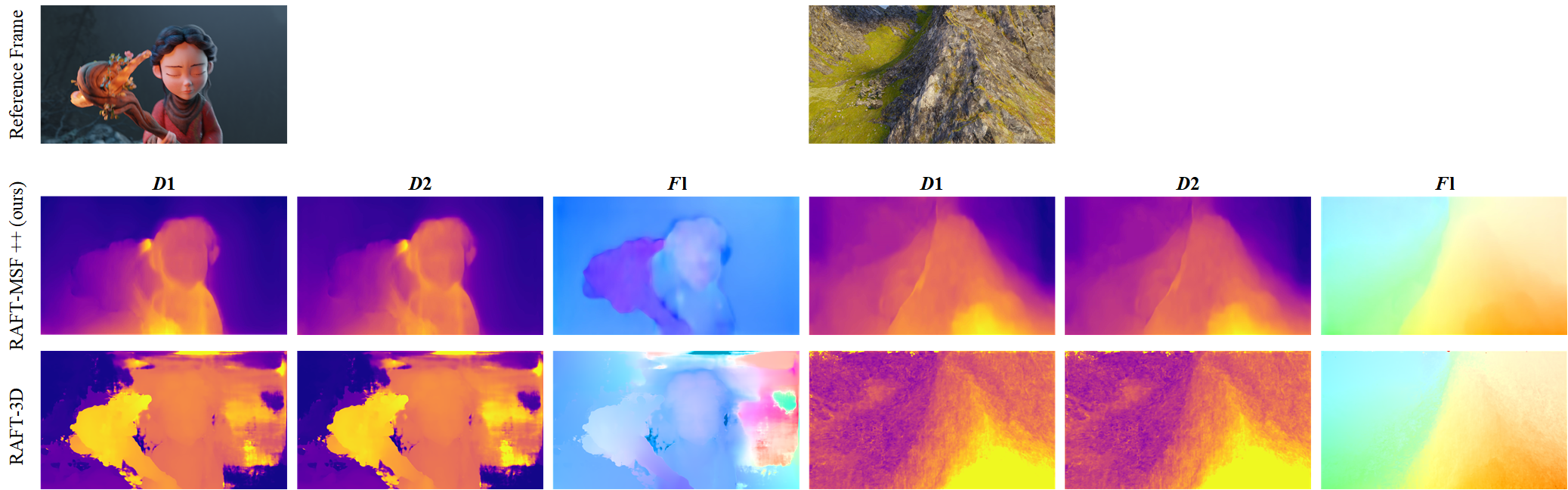}
	\caption{Zero-shot generalization results on the Spring dataset~\cite{mehl2023spring} compared with RAFT-3D~\cite{teed2021raft}. RAFT-3D is a supervised method that takes stereo image pairs as input, while our RAFT-MSF++ is trained in a self-supervised manner and performs inference using only monocular sequences. Despite weaker supervision and fewer inputs, our model yields competitive scene flow estimates.
	}
	
	\label{fig:generalization_spring}
\end{figure*}

\section{Generalization Ability}

To evaluate our model’s robustness to distribution shifts, we perform zero-shot cross-dataset testing on Cityscapes~\cite{cordts2016cityscapes}, nuScenes~\cite{caesar2020nuscenes}, and high-resolution Spring~\cite{mehl2023spring}. All evaluations use our model trained solely on KITTI~\cite{geiger2013vision} with no fine-tuning.

As illustrated in Fig.~\ref{fig:generalization}, our method demonstrates remarkable generalization potential across diverse unseen environments. On Cityscapes (left), the model accurately captures pedestrian motion, a category that is underrepresented in the KITTI training set, showcasing its ability to handle complex dynamic objects. On nuScenes (right), the framework maintains robustness across varying camera configurations and lighting conditions, producing stable estimates even in challenging nighttime scenarios. 

Furthermore, we provide a qualitative comparison with the state-of-the-art supervised stereo method RAFT-3D~\cite{teed2021raft} on the Spring dataset in Fig.~\ref{fig:generalization_spring}. It is worth noting that zero-shot quantitative evaluation on Spring for monocular methods inherently suffers from severe scale ambiguities due to the drastically different camera intrinsics and resolutions compared to KITTI. However, from a qualitative perspective, our self-supervised monocular framework achieves visual quality comparable to the supervised stereo RAFT-3D, effectively recovering the underlying structural layouts and relative motion boundaries of complex rendered scenes. These visual results suggest that our GMF-based multi-frame fusion captures meaningful spatial-temporal representations that generalize reasonably well to unseen domains.

\section{Limitations}\label{E-3}   % Latency and Limitations
While RAFT-MSF++ achieves strong results in self-supervised monocular scene flow estimation, it has two main limitations. First, it requires stereo images for training and assumes known camera intrinsics, which we plan to address with a self-calibration mechanism.

Second, RAFT-MSF++ underperforms in scenes with small ego-motion, as also observed in prior work~\cite{hur2020self,hur2021self,bayramli2023raft}. 
The primary cause is the relatively large disparity estimation error (D1 Error) in these scenarios, as illustrated in Fig.~\ref{fig:limit}.

Specifically, scene flow networks for joint disparity and scene flow estimation implicitly use stereo cues from consecutive frames. Small ego-motion reduces the baseline, weakening these cues and worsening depth estimation in static or low-motion scenes.

In contrast, EMR-MSF~\cite{jiang2023emr} avoids this issue by decoupling monocular depth estimation (SDFA-Net~\cite{zhou2022self}) and motion field prediction (RAFT-3D~\cite{teed2021raft}), and by leveraging an ego-motion rigidity prior. In future work, we plan to integrate these strengths into our pipeline to overcome the current limitations.

\section{Additional Results}
\begin{figure*}[h]
	\centering
	\includegraphics{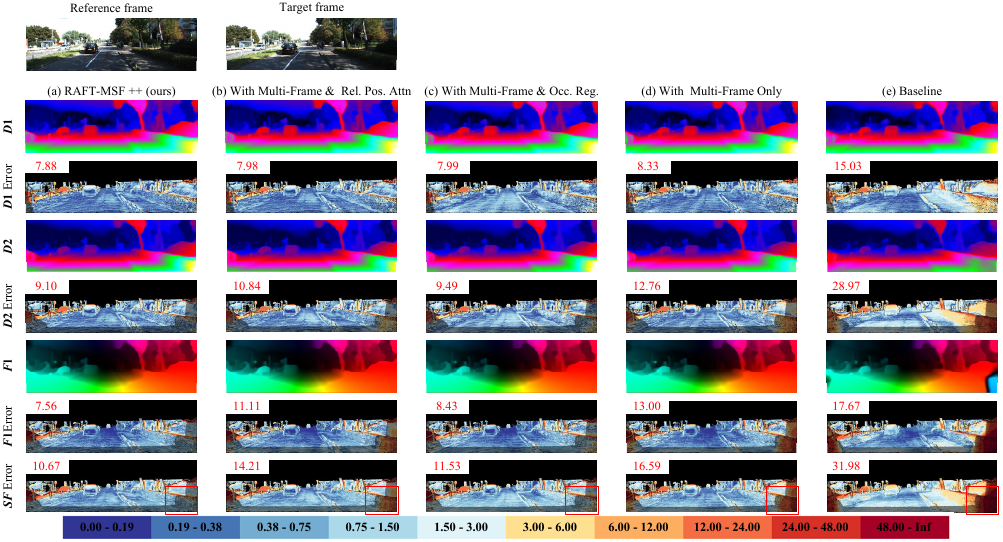}
	\caption{Qualitative results from the ablation study illustrating the impact of our main contributions. 
	%	The top row shows the reference frame and the forward target frame. 
		Columns (a)–(e) depict results from different model variants: (a) the full model, (b) with multi-frame input and relative positional attention, (c) with multi-frame input and occlusion regularization, (d) with multi-frame input only, and (e) the baseline without either component. Each row presents disparity maps and their error maps for the reference frame (D1/D1 Error), warped disparity maps and error maps for the target frame (D2/D2 Error), optical flow and its error maps (Fl/Fl Error), and scene flow error maps (SF Error). In the error maps, correct estimations are indicated in \colorbox{blue}{\textcolor{white}{blue}}, incorrect estimations in \colorbox{red}{\textcolor{white}{red}}, and occluded pixels are shown as dark regions. Outlier rates are overlaid on the corresponding error maps.}

	\label{fig:ablation_main}
\end{figure*}

We provide qualitative results for the ablation study on the main contributions. Fig.~\ref{fig:ablation_main} compares the full RAFT-MSF++ model with several ablated variants: (a) full model, (b) w/ multi-frame + relative positional attention, (c) w/ multi-frame + occlusion regularization, (d) w/ multi-frame only, and (e) baseline without these components.

The visual results show that the full model produces significantly fewer errors across disparity, optical flow, and scene flow, particularly in occluded regions and areas with large motion. The overlaid outlier rates on the error maps further highlight the contribution of each module to overall performance.

%% file: sec/abla05.tex
%\vspace*{-1.2em}
\begin{table}[ht]
	\caption{Performance of Multi-Mono-SF under different input frame settings.}
	\label{tab5}%
	\centering
	\fontsize{9pt}{11pt}\selectfont
%	\begin{tabular}{@{}p{2.5cm}cccc@{}}
	\begin{tabular}{lcccc}
		\toprule
		Setting & D1-all & D2-all & Fl-all & SF-all \\
		\midrule
		Reported (3-frame) & 32.87 & 34.70 & 22.75 & 46.15 \\
		Reported (5-frame) & \textbf{27.33} & \textbf{30.44} & 18.92 & 39.82 \\
		zeroed states (3-frame) & 27.34 & 30.62 & \textbf{18.81} & \textbf{39.81} \\
		\bottomrule
	\end{tabular}
	
\end{table}
%\vspace*{-2.0em}